\PassOptionsToPackage{table,xcdraw}{xcolor}
\documentclass[runningheads]{llncs}
\usepackage[T1]{fontenc}
\usepackage{graphicx,verbatim}
\usepackage{xcolor} 
\usepackage{amsfonts}
\usepackage{algorithm}
\usepackage{algorithmic}
\usepackage{textcomp}
\usepackage{wrapfig}
\usepackage{svg}
\usepackage{amssymb}
\usepackage{amsmath}
\usepackage{multirow}
\usepackage{adjustbox}
\usepackage{url}
\usepackage{booktabs}       
\usepackage{hyperref}
\usepackage{xpatch}

\makeatletter
\ExplSyntaxOn
\cs_new:Npn \bibColoredItems #1#2
  {
    \clist_map_inline:nn {#2} { \cs_new:cpn {bib@colored@##1} {#1} } 
  }
\ExplSyntaxOff

\newcommand\bib@setcolor[1]{%
  \ifcsname bib@colored@#1\endcsname
    \expanded{\noexpand\color{\csname bib@colored@#1\endcsname}}%
  \else
    \normalcolor
  \fi
}

\IfPackageLoadedTF{hyperref}{\@tempswatrue}{\@tempswafalse}
\if@tempswa
  \xpatchcmd\@bibitem {\H@item}{\bib@setcolor{#1}\H@item}{}{\PatchFailed}
  \xpatchcmd\@lbibitem{\H@item}{\bib@setcolor{#2}\H@item}{}{\PatchFailed}
\else
  \xpatchcmd\@bibitem {\item}  {\bib@setcolor{#1}\item}  {}{\PatchFailed}
  \xpatchcmd\@lbibitem{\item}  {\bib@setcolor{#2}\item}  {}{\PatchFailed}
\fi
\makeatother

\definecolor{revisioncolor}{HTML}{000000} 

\newcommand{\revision}[1]{\textcolor{revisioncolor}{#1}}  

\definecolor{modifycolor}{HTML}{000000} 

\begin{document}
\title{PHGNN: A Novel Prompted Hypergraph Neural
Network to Diagnose Alzheimer's Disease}
\titlerunning{A Prompted Hypergraph Neural
Network to Diagnose AD}
%

\author{Chenyu Liu\inst{1} \and
 Luca Rossi \inst{1,2} }
\authorrunning{C. Liu et al.}
%
\institute{Department of Electrical and Electronic Engineering, The Hong Kong Polytechnic University, Hong Kong
SAR, China
\and Department of Data Science and Artificial Intelligence The Hong Kong Polytechnic University, Hong Kong
SAR, China \\
\email{luca.rossi@polyu.edu.hk}}
%
\maketitle              

\begin{abstract}
The accurate diagnosis of Alzheimer's disease (AD) and prognosis of mild cognitive impairment (MCI) conversion are crucial for early intervention. However, existing multimodal methods face several challenges, from the heterogeneity of input data, to underexplored modality interactions, missing data due to patient dropouts, and limited data caused by the time-consuming and costly data collection process.
In this paper, we propose a novel Prompted Hypergraph Neural Network (PHGNN) framework that addresses these limitations by integrating hypergraph based learning with prompt learning. Hypergraphs capture higher-order relationships between different modalities, while our prompt learning approach for hypergraphs, adapted from NLP, enables efficient training with limited data. Our model is validated through extensive experiments on the ADNI dataset, outperforming SOTA methods in both AD diagnosis and the prediction of MCI conversion.

\keywords{Mild cognitive impairment \and Alzheimer's disease,
Multimodal learning \and Hypergraph neural networks \and Prompt learning.}

\end{abstract}
\section{Introduction}
Alzheimer’s disease (AD) is one of the most common diseases in elderly people, caused by the irreversible loss of neurons and genetically complex disorders. Therefore, accurate recognition of AD and its precursor stage, mild cognitive impairment (MCI), has attracted widespread attention, as MCI has been shown to be the optimal stage to treat in order to prevent the MCI-to-AD conversion \cite{spasov2019parameter}. This in turn highlights the importance of progressive MCI prediction, which aims to distinguish progressive MCI (pMCI), which may progress to AD within 36 months, from stable MCI (sMCI). \revision{In recent years, several deep learning based computer-aided diagnosis methods \cite{polsterl2021combining,gao2021task,yu2024transformer,zhang2024modality,xu2025domain} have been proposed to diagnose AD and MCI using imaging and non-imaging data as different modalities that carry complementary information about the disease.}

However, existing multimodal approaches for AD diagnosis suffer from three major limitations: 1) different modalities are highly heterogeneous, especially between imaging and non-imaging data; 2) patient dropouts can result in some subjects not having data from a specific modality, resulting in missing data; 3) more in general, in the field of clinical research the amount of data is often very limited, hindering the performance of deep learning models that need to be trained on large amounts of data in order to make accurate predictions.

The objective of the present study is to address these limitations. To tackle the first issue, \revision{we propose to use a hypergraph neural network (HGNN) based approach, which has already shown its effectiveness in AD diagnosis \cite{shao2020hypergraph,zuo2021multimodal,aviles2022multi}, based on the following two intuitions:}
1) hypergraphs can model higher-order relationships directly by allowing a single hyperedge to connect multiple nodes, thus allowing us to better represent the rich interactions between different modalities \cite{cai2022hypergraph};
2) hypergraphs can leverage the remaining modalities by exploiting the connections between the available data points. This ability to integrate information from multiple modalities simultaneously makes hypergraphs more resilient to missing data compared to traditional graphs.

A promising solution to the issue of data availability (e.g., due to patient dropouts) is prompt learning \cite{min2023recent}.
However, adapting prompt learning to hypergraph based disease prediction is a non-trivial problem. In NLP, the input is just a sequence of words while hypergraphs are fundamentally more complex as they incorporate both feature and structural information.
Therefore, one needs to reformulate language prompts as hypergraph prompts. 

To tackle the above issues, we make the following contributions: (1) we propose PHGNN, a framework for AD diagnosis and MCI conversion prognosis which uses MRI, PET, and non-imaging clinical data for hypergraph pretraining followed by prompt tuning; (2) we extend the concept of prompt learning, originally introduced in the context of NLP, to HGNNs, enabling us to efficiently train our model with limited data; 
(3) \revision{we conduct extensive experiments comparing PHGNN against other multimodal approaches and tuning strategies.}

\section{ Prompted Hypergraph Neural Network}\label{sec:method}
\begin{figure*}[t!]
    \centering
    \includegraphics[width=\textwidth]{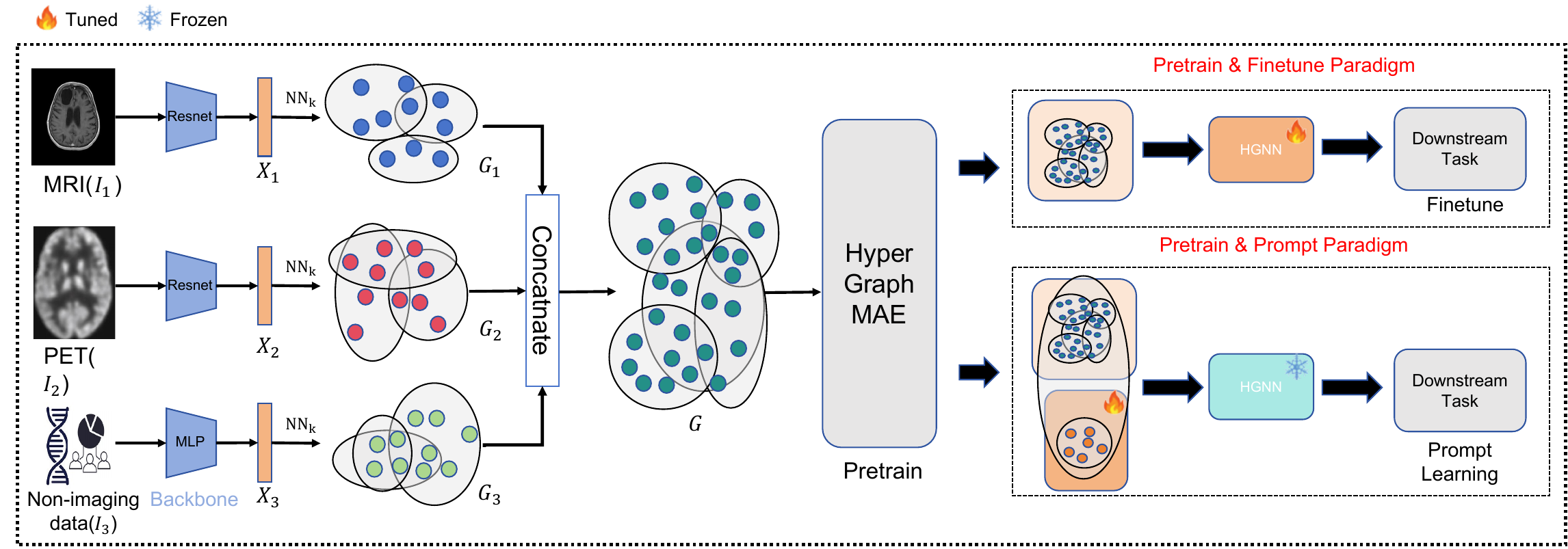} 
    \caption{
    We extract features from the available modalities and use $k$-NN to compute the corresponding hypergraphs. These are then concatenated to create the hypergraph $G$. $G$ is first used for pre-training using a HyperGraphMAE. Then the pre-trained HGNN encoder is used for tuning on the downstream task. In the Pretrain \& Finetune Paradigm, the HGNN is directly optimized through the hypergraph $G$. In the Pretrain \& Prompt Paradigm, the HGNN is frozen and we construct a learnable prompt sub-hypergraph $G_p$ which is combined with $G$ to generate a manipulated hypergraph $G_m$. In this case only the learnable prompt tokens in $G_p$ are optimized during finetuning.
    }
    \label{fig:framework}
\end{figure*}

PHGNN is a semi-supervised prompted hypergraph learning framework composed of 3 key components: 1) modality-aware hypergraph representation, 2) self-supervised hypergraph pretraining, and 3) hypergraph prompt learning. 
Figure~\ref{fig:framework} shows an overview of the structure of the proposed framework.

\subsection{Hypergraph representation}
\label{subsec:hypergraph}
\revision{Unlike graphs, where edges connect two nodes, the hyperedges of a hypergraph connect multiple nodes, enabling the representation of higher-order relationships between node subsets with shared features which in turn allows us to model complex interactions among the input modalities.} In particular, we create an undirected node-attributed hypergraph $G = (V, E, H)$, with node set $V$, hyperedge set $E$, and adjacency matrix $H \in \mathbb{R}^{|V| \times |E|}$. Each node $v \in V$ corresponds to a subject (i.e., a patient) and each hyperedge $e \in E$ connects a subset of nodes. For each node $v$, we create a hyperedge $e \in E$ by connecting $v$ to its $k$-nearest neighbors based on the Euclidean distance in the node feature space. Then the adjacency matrix is $H(v,e)= 1$ if $v \in e$, or $H(v,e)= 0$ otherwise.

Recall that the input data is composed $m$ modalities $I =
\{I_1,I_2,...,I_m\}$. For each modality we feed the input data to the feature extraction backbone to obtain $m$ sets of node features $X =
\{X_1,X_2,... ,X_m\}$, where $X_i \in \mathbb{R}^{|V| \times d}$ and $d$ is the dimension of the input features. For each modality $X_i =
\{x_i^1,x_i^2,... ,x_i^{|V|}\}$, we build a hypergraph $G_i$ by constructing the corresponding set of hyperedges $E_i$ using the $k$-nearest neighbour ($k$-NN) method, as discussed above. For each modality, this results in \( |V| \) hyperedges, each linking \( k + 1 \) nodes. Consequently, we obtain an incidence matrix \( H_m \in \mathbb{R}^{|V| \times |V|} \), where \( |V| \times (k + 1) \) entries are set to 1, and all other entries are 0. The resulting hypergraphs from each modality are finally concatenated to obtain the hypergraph $G = \{G_1 || G_2 || ... || G_m\}$ using the \emph{coequal fusion mechanism} introduced in \cite{cai2022hypergraph}.

\begin{algorithm}
\caption{Overall Prompt Learning Process}
\label{alg:training}
\begin{algorithmic}[1]
    \STATE {\bf Input:} hypergraph $G$, pre-trained HGNN model $h_{\theta^*}$ with frozen $\theta^*$, prompt tokens $P$ with prompt parameter $\theta$, number of epochs $\mathcal{E}$, learning rate $\eta$, training mask $M_T$, validation mask $M_V$
    \STATE {\bf Output:} Optimal prompt tokens $P$ with prompt parameter $\theta$
    \STATE Initialize prompt parameters $\theta$
    \FOR{epoch $= 1$ to $\mathcal{E}$}
        \STATE Construct the prompt sub-hypergraph $G_p$
        \STATE $G_m = CONCATE(G, G_p)$
        \STATE Compute predictions $\hat{y} = h_{\theta^*}(G_m)$
        \STATE Compute loss $\mathcal{L}_{sup}(\hat{y}[M_T], y[M_T])$
        \STATE Backpropagate and update parameters $\theta \leftarrow \theta - \eta \cdot \nabla_\theta \mathcal{L}_{sup}$
        \STATE Compute predictions $\hat{y} = h_{\theta^*}(G_m)$
        \STATE Evaluate model on validation data \\
        $Evaluate(\hat{y}[M_V], y[M_V])$
        \STATE Save model if validation performance improves
    \ENDFOR
    \STATE {\bf Return:} $P$ with $\theta$ 
\end{algorithmic}
\end{algorithm}

\subsection{HyperGraphMAE pretraining}
\revision{Inspired by GraphMAE \cite{hou2022graphmae}, we propose HyperGraphMAE}, a generative self-supervised learning framework that we use to pretrain the HGNN at the core of our pipeline. Given a hypergraph $G = (V, E, H)$, let \(h_E\) be a hypergraph encoder, \(h_D\) a hypergraph decoder, and $Z \in \mathbb{R}^{|V| \times {d_z}}$ the corresponding hidden state, i.e.,
\begin{equation}\label{eq:hgmae}
Z = h_E(G, X), \quad G' = h_D(G, Z)\,,
\end{equation}
where $G'$ is the reconstructed hypergraph, and both \(h_E\) and \(h_D\) are expressive HGNNs capable of leveraging information from the node neighbourhood.

HyperGraphMAE is trained to reconstruct masked node features of the hypergraph $G$ from subsection \ref{subsec:hypergraph}. To this end, during the encoding phase a subset of nodes \(V_m \subset V\) of $G$ is randomly selected and their features are masked with a learnable token \(x_{[M]}\). After the encoding, the embeddings of the masked nodes are re-masked with another token \(z_{[M]}\) before being passed to \(h_D\). This encourages the decoder to reconstruct node features using information from neighboring nodes. The goal is then to reconstruct the masked node features \(\hat{X}\) given the partially observed node features \(X\) and the adjacency matrix \(H\) of $G$. We use the Scaled Cosine Error (SCE) as the reconstruction loss, which has the advantage of reducing the sensitivity and selectivity issues of the mean squared error, i.e.,
\[
L_{\text{SCE}} = \frac{1}{|V_m|} \sum_{v_i \in V_m} \left( 1 - \frac{x_i^\top x'_i}{\|x_i\| \|x'_i\|} \right)^\gamma \,,
\]
where \(x_i\) and \(x'_i\) are the original and reconstructed node features, respectively, and \(\gamma \geq 1\) is a scaling factor that improves selectivity by down-weighting the contribution of easy samples during training.

\subsection{Hypergraph prompt}
\revision{In NLP, a prompt is a sequence of tokens added to the input sentence. To adapt this concept to hypergraphs, we need to define three key elements: (1) prompt token, (2) prompt structure, and (3) pattern insertion.}

\noindent\paragraph{\normalfont \textbf{Prompt tokens.}}
Given a hypergraph $G = (V,E, H)$, we introduce the prompt sub-hypergraph $G_p = (P, E_p, H_p)$, where $P = \{p_1, p_2, ..., p_{|P|}\}$ represents the set of \( |P| \) learnable prompt tokens and $E_p$ is the hyperedge set inside the prompt sub-hypergraph. Each token \( p_i \in P \) is represented as a token vector \( p_i \in \mathbb{R}^{1 \times d} \), matching the dimensions of the node features in the input hypergraph. \revision{In practice, \( |P| \) is much smaller than \( N \) and \( |P| \ll d_z \), where \( d_z \) is the hidden layer size in the pre-trained hypergraph model. 
}

\noindent\paragraph{\normalfont \textbf{Prompt structure.}}
we model the relationship among prompt tokens using the $k$-NN method based on the Euclidean distance between the tokens. This results in an incidence matrix $H_p$ for the prompt sub-hypergraph $G_p$. Note that $H_p$ is recomputed at the start of every training epoch given the latest prompts $P$.
\renewcommand{\arraystretch}{1.5} 
\begin{table*}[t]\centering
\caption{Classification performance on AD vs CN and pMCI vs sMCI. For each metric we show the average ($\pm$ std deviation) over 5 folds (best model highlighted in green).}
\fontsize{8}{8}\selectfont 
\begin{tabular}{l|c c c c|c c c c}
\hline
\multirow{2}{*}{} & \multicolumn{4}{c|}{\textbf{AD vs CN}} & \multicolumn{4}{c}{\textbf{pMCI vs sMCI}} \\ \cline{2-9} 
 & \textbf{BACC} & \textbf{SEN} & \textbf{SPE} & \textbf{AUC} & \textbf{BACC} & \textbf{SEN} & \textbf{SPE} & \textbf{AUC} \\ \hline
GNNs \cite{parisot2018disease} & 86.8$\pm$0.3 & 89.7$\pm$0.2 & 83.7$\pm$0.4 & 91.5$\pm$0.4 & 70.9$\pm$0.7 & 52.7$\pm$0.7 & \cellcolor[HTML]{CAFFCA}89.1$\pm$0.6 & 75.9$\pm$0.5 \\ \hline
HGNN \cite{feng2019hypergraph}  & 89.7$\pm$1.3 & 88.7$\pm$1.0 & 90.7$\pm$1.2 & 94.1$\pm$0.8 & 74.7$\pm$0.7 & 72.3$\pm$0.9 & 77.1$\pm$0.4 & 77.7$\pm$0.8 \\ \hline
HGNN+ \cite{gao2022hgnn+} & 89.4$\pm$0.9 & \cellcolor[HTML]{CAFFCA}91.6$\pm$1.1 & 87.1$\pm$0.9 & 94.2$\pm$1.2 & 74.6$\pm$1.3 & \cellcolor[HTML]{CAFFCA}77.5$\pm$1.2 & 71.8$\pm$1.3 & 77.3$\pm$1.6 \\ \hline
PT-DCN \cite{gao2021task} & 92.7 & 91.7 & 93.5 & 96.4 & 75.3 & 70.8 & 78.4 & 77.8 \\ \hline
SPDN \cite{xu2025domain} & 92.9 & 91.9 & 93.6 & 96.6 & 76.2 & 62.8 & 80.6 & 77.3 \\ \hline
MFF \cite{zhang2024modality} & 91.2$\pm$0.5 & 91.5$\pm$1.5 & 90.9$\pm$0.9 & 95.5$\pm$0.4 & 77.6$\pm$1.1 & 74.9$\pm$1.6 & 80.2$\pm$0.6 & 80.6$\pm$0.5 \\ \hline
PHGNN & \cellcolor[HTML]{CAFFCA}93.2$\pm$0.6 & 90.6$\pm$1.9 & \cellcolor[HTML]{CAFFCA}95.6$\pm$0.6 & \cellcolor[HTML]{CAFFCA}97.2$\pm$1.2 & \cellcolor[HTML]{CAFFCA}79.6$\pm$0.6 & 75.3$\pm$0.8 & 83.8$\pm$1.5 & \cellcolor[HTML]{CAFFCA}82.8$\pm$1.1 \\ \hline
\end{tabular}
\label{tab:comparison}
\end{table*}
\noindent\paragraph{\normalfont \textbf{Pattern insertion.}}
Let $\psi$ represent the insertion method to add the prompt graph $G_p$ to the input hypergraph $G$, resulting in the manipulated hypergraph $G_m = \psi(G, G_p)$. 
We leverage the fact that a hyperedge can connect multiple nodes and for each prompt token we define a hyperedge connecting it to all the nodes of $G$, i.e., $E_m = E \cup E_p \cup  \Bigl\{ \lbrace  v_1,v_2,...,v_N,p_i \rbrace, \forall p_i \in P \Bigr\}$,
where $E_m$, $E$, and $E_p$ are the sets of hyperedges of $G_m$, $G$, and $G_p$, respectively, and $\lbrace  v_1,v_2,...,v_N,p_i \rbrace$ denotes a new hyperedge connecting the node corresponding to the token $p_i$ of $G_p$ to the $N$ nodes of $G$.

\subsection{Overview of the prompt learning process}
The main objective of our framework is a binary class prediction task (AD vs CN, sMCI vs pMCI). As shown in Algorithm \ref{alg:training}, the overall prompt learning process starts with a hypergraph $G$ (see \ref{subsec:hypergraph}) and a pre-trained HGNN model $h_\theta$, where the pre-trained weights are from the HyperGraphMAE encoder $h_E$ of Eq.~\ref{eq:hgmae}. To split the data into train and validation sets, we define the training mask $M_T$ and the validation mask $M_V$, which are binary masks used to control which nodes of the graph are used during different phases of the training process. 

Given this setting, a sub-hypergraph $G_p$ with a set of prompt tokens is initialized and constructed. Then $G$ and $G_p$ are concatenated through hyperedges to produce the hypergraph $G_m$, which is processed by the frozen pre-trained HGNN model $h_\theta$ and only the learnable prompt tokens $P$ are optimized. Note that for each epoch, $G_p$ is reconstructed again with the updated set of tokens $P$. The resulting output is a probability distribution over the target classes. We use the supervised cross-entropy (CE) loss $\mathcal{L}_{sup} = \sum_{i=1}^{N} CE(\hat{y}_i, y_i)\label{eq:supervised}$, where \( \hat{y}_i \) represents the output probability and \( y_i \) corresponds to the ground truth label for the \( i \)-th node (i.e., the $i$-th patient).

\section{Experimental results}\label{sec:experiments}
\textbf{Data description and implementation details.} 
We evaluate our framework on the Alzheimer's Disease Neuroimaging Initiative (ADNI) \cite{jack2008alzheimer}. We trained the network on the ADNI-1 dataset and evaluated its performance using the ADNI-2 dataset. The subjects in the datasets are categorized following \cite{gao2021task} into 4 groups: AD, cognitively normal (CN), pMCI, and sMCI. \revision{We use three modalities: MRI, PET, and tabular data. For MRI and PET images, we adopt the same pre-processing procedure as \cite{gao2021task}. For the tabular data, we use the same set of variables and pre-processing principles as \cite{jarrett2019dynamic}. For ADNI1, we further split the data using a 5-fold cross-validation strategy. }\revision{In addition, pretraining and finetuning are performed over the same training subjects.}
We use a ResNet-50 to extract features from MRI and PET images and a multi-layer perceptron to extract features from the tabular data. Given these features, the hypergraph representing the set of patients is constructed by setting the $k$-NN parameter to 30 (see subsection \ref{subsec:hypergraph}). We optimize the model using \textsc{AdamW} with a weight decay of 1e-4 and a learning rate of 3e-4. We experimentally tuned the value of $|P|$ as discussed in the ablation study. For HyperGraphMAE, we use a uniform random sampling strategy without replacement, with 75\% of the nodes being masked. We set $\gamma$ in $L_{\text{SCE}}$ be 2. Our code can be accessed at \url{https://anonymous.4open.science/r/PHGNN-B3B3/}.\\

\noindent\textbf{Evaluation protocol.}
We evaluate our model (PHGNN) on multimodal classification tasks, specifically AD vs CN and pMCI vs sMCI, against SOTA alternatives as well as other graph \revision{parameter-efficient fine-tuning (PEFT) methods}. To assess the performance of the proposed method, we calculate four metrics with their average and standard error over the three folds: balanced accuracy (BACC), sensitivity (SEN), specificity (SPE), and area under the curve (AUC).\\

\noindent\textbf{Comparison with existing methods.}
We compare the classification performance of PHGNN with that of three SOTA multimodal classification models (PT-GCN \cite{gao2021task}, \revision{SPDN \cite{xu2025domain}, and Modality-Flexible Framework (MFF) \cite{zhang2024modality}), two hypergraph based models based (HGNN \cite{feng2019hypergraph} and HGNN+ \cite{gao2022hgnn+}), and a GNN based method \cite{parisot2018disease}. The hypergraph based models mirror our pipeline without resorting to prompt tuning.
The GNN, HGNN, and HGNN+ are also pre-trained using the HyperGraphMAE method for fair comparison. We run MFF using publicly available code, while we use the reported results for PT-GCN and SPDN as they are based on exactly the same ADNI subjects and data split setting we use.} Table \ref{tab:comparison} shows that PHGNN outperforms the other models, achieving the highest BACC and AUC in both tasks, showing its ability to handle complex classification problems in clinical settings. In AD vs CN, PHGNN achieves the highest performance among all models, with a BACC of 0.932 and an AUC of 0.9721.
In the more challenging task of pMCI vs sMCI, PHGNN again shows superior performance with a BACC of 0.7962 and an AUC of 0.8278, outperforming \revision{MFF} across all metrics and by 2.07\% in BACC and 2.23\% in AUC. Note that we use HGNN and not HGNN+ in PHGNN as the former has higher AUC in this task.\\
\begin{table*}[t!]
\centering
\caption{Prompt learning strategies on AD vs CN and pMCI vs sMCI. For each metric we show the average ($\pm$ std deviation) over 5 folds (best model highlighted in green).}
\fontsize{8}{8}\selectfont 
\begin{tabular}{c|c c c c|c c c c}
\hline
\multirow{2}{*}{} & \multicolumn{4}{c|}{\textbf{AD vs CN}} & \multicolumn{4}{c}{\textbf{pMCI vs sMCI}} \\ \cline{2-9} 
 & \textbf{BACC} & \textbf{SEN} & \textbf{SPE} & \textbf{AUC} & \textbf{BACC} & \textbf{SEN} & \textbf{SPE} & \textbf{AUC} \\ \hline
Finetune  & 89.7$\pm$1.3 & 88.7$\pm$1.0 & 90.7$\pm$1.2 & 94.1$\pm$0.8 & 74.7$\pm$0.7 & 72.3$\pm$0.9 & 77.1$\pm$0.4 & 77.7$\pm$0.8 \\ \hline
GPF \cite{fang2024universal} & 90.3$\pm$1.9 & 87.4$\pm$1.1 & 93.0$\pm$1.5 & 95.3$\pm$1.4 & 75.7$\pm$0.9 & 75.3$\pm$1.1 & 76.0$\pm$0.9 & 80.2$\pm$0.8 \\ \hline
GPF-Plus \cite{fang2024universal} & 89.0$\pm$1.4 & 89.7$\pm$1.2 & 88.4$\pm$1.6 & 93.3$\pm$1.5 & 71.1$\pm$0.5 & 47.6$\pm$0.4 & \cellcolor[HTML]{CAFFCA}94.5$\pm$0.8 & 79.0$\pm$1.8 \\ \hline
LoRA \cite{hu2022lora} & 89.0$\pm$0.1 & 86.1$\pm$0.3 & 91.8$\pm$0.2 & 93.0$\pm$0.1 & 76.2$\pm$0.5 & 70.7$\pm$0.7 & 81.7$\pm$0.5 & 79.5$\pm$1.0 \\ \hline
 AdaGNN\cite{li2024adaptergnn} & 90.3$\pm$4.0 & 86.1$\pm$1.3 & 94.5$\pm$0.6 & 94.5$\pm$0.1 & 77.6$\pm$0.6 & 69.2$\pm$1.4 & 86.0$\pm$0.5 & 80.5$\pm$0.2 \\ \hline
PHGNN & \cellcolor[HTML]{CAFFCA}93.2$\pm$0.6 & \cellcolor[HTML]{CAFFCA}90.6$\pm$1.9 & \cellcolor[HTML]{CAFFCA}95.6$\pm$0.6 & \cellcolor[HTML]{CAFFCA}97.2$\pm$1.2 & \cellcolor[HTML]{CAFFCA}79.6$\pm$0.6 & \cellcolor[HTML]{CAFFCA}75.3$\pm$0.8 & 83.8$\pm$1.5 & \cellcolor[HTML]{CAFFCA}82.7$\pm$1.1 \\ \hline
\end{tabular}
\label{tab:prompt comparison}
\end{table*}

\noindent\textbf{Comparison with other PEFT methods.}
To evaluate the effectiveness of our prompting strategy, we compare it with two other prompt learning methods that can be directly used with hypergraphs, GPF and GPF-Plus \cite{fang2024universal}, \revision{and two adapter methods LoRA \cite{hu2022lora} and AdapterGNN \cite{li2024adaptergnn}}, as well as the Pre-train \& Finetune approach. GPF and GPF-plus primarily incorporate soft prompts into all node features of the input graph. The results are shown in Table \ref{tab:prompt comparison}.
In the AD vs CN task, PHGNN achieves the best performance across all metrics. Compared to the traditional Pre-train \& Finetune approach, which obtains a BACC of 0.8971 and an AUC of 0.9419, PHGNN improves both classification accuracy and the balance between sensitivity and specificity. When compared to the other PEFT methods, PHGNN also shows superior performance, achieving a BACC of 0.7962 and an AUC of 0.8278 in pMCI vs sMCI.\\

\noindent\textbf{Ablation study.}
We then perform an ablation study on the sMCI vs pMCI task in order to evaluate the effect of 1) varying the number of prompt tokens $|P|$; 2) replacing the HGNN with a GNN denoted as Prompt GNN; 3) replacing our \emph{Prompt as hypergraph} strategy, where a hypergraph prompt has multiple prompt tokens and a non-trivial structure, with \textit{Prompt as token}, where we treat prompt tokens as independent prompts without considering the inner structure (denoted as PHGNN w/o S). Table~\ref{tab:auc_comparison} (Right) shows the results of this ablation study. Firstly, we observe that the number of tokens $|P|$ can have a significant influence on the result of the prompt learning, which is in accordance with what is highlighted in~\cite{wu2023survey}. Therefore, to get optimal results it is important to select the optimal number of tokens $|P|$. In future work, one possibility would be to leverage meta-learning to automatically select the best value of $|P|$. Secondly, although the prompt learning strategy leads to an improvement over the original GNN performance (see Table~\ref{tab:comparison}), this is still inferior when compared to PHGNN. The results clearly demonstrate the importance of the inner structure of prompt graphs, as opposed to simply treating the tokens as independent prompts.\\
\begin{table*}[t]
\fontsize{8}{8}\selectfont 
\caption{(\textbf{Left}) Using diferent combinations of multimodal data on pMCI vs sMCI. (\textbf{Right}) AUC scores for various settings on sMCI vs pMCI (best model in green).}
\centering 
\begin{tabular}{c|c|c|c|c|c|c}
      \toprule
      \textbf{MRI} & \textbf{PET} & \textbf{Tabular} & \textbf{BACC} & \textbf{SEN} & \textbf{SPE} & \textbf{AUC} \\
      \midrule
      \checkmark &  &  & 72.9 & 66.3 & 79.5 & 77.1 \\
      & \checkmark &  & 74.0 & 67.2 & 80.8 & 77.4 \\
      &  & \checkmark & 72.4 & 65.1 & 79.6 & 76.6 \\
      \checkmark & \checkmark &  & 76.0 & 67.2 & 83.9 & 79.6 \\
      \checkmark &  & \checkmark & 73.8 & 69.2 & 78.5 & 78.9 \\
      & \checkmark & \checkmark & 77.4 & 72.0 & 82.8 & 80.4 \\
      \checkmark & \checkmark & \checkmark & \cellcolor[HTML]{CAFFCA}79.6 & \cellcolor[HTML]{CAFFCA}75.3 & \cellcolor[HTML]{CAFFCA}83.8 & \cellcolor[HTML]{CAFFCA}82.7 \\
      \bottomrule
    \end{tabular}    
\hfill
\begin{tabular}{c|c c c c}
      \hline
      \multirow{2}{*}{} & \multicolumn{4}{c}{\textbf{AUC}} \\ \cline{2-5}
      \textbf{$|P|$} & \textbf{8} & \textbf{16} & \textbf{32} & \textbf{64} \\ \hline
      Prompt GNN & - & 77.4 & - & - \\ \hline
      PHGNN w/o S & - & 79.8 & - & - \\ \hline
      PHGNN & 76.8 & \cellcolor[HTML]{CAFFCA}82.7 & 81.4 & 72.8 \\ \hline
    \end{tabular}
    \label{tab:auc_comparison}
    \label{tab_modal}
\end{table*}

\noindent\textbf{Contribution of different modalities.}
\revision{We evaluate the contributions of different data modalities in PHGNN, as shown in Table \ref{tab_modal} (Left). The first three rows represent diagnostic results for MCI prediction using each modality independently. PET scans outperform sMRI scans, with PET achieving a BACC of 74.01\% and AUC of 77.4\%, surpassing sMRI by 1.04\% in BACC and 0.3\% in AUC. This aligns with prior research \cite{samper2018reproducible,ota2015effects}, as hypometabolism in PET scans appears earlier in the disease than atrophy in sMRI \cite{jack2010hypothetical}. Non-imaging clinical data perform worse, with a BACC and AUC 0.57\% and 0.43\% lower than the lowest imaging modality.
The next three rows show the results of modality fusion. Adding a modality boosts performance, demonstrating the importance of multimodal fusion and PHGNN's ability to handle it. Combining all three modalities yields the best diagnostic performance, suggesting that features from different modalities provide complementary insights, improving classification accuracy.}\\

\noindent\textbf{Parameter efficiency analysis.}
As a final experiment, we compare the number of tunable parameters of the proposed PHGNN model \revision{and other tuning strategies} in the sMCI vs pMCI classification task: traditional fine-tuning ($\sim0.5$M), GPF~\cite{fang2024universal} ($\sim5$K), GPF-plus~\cite{fang2024universal} ($\sim0.21$M), LoRA~\cite{hu2022lora} ($\sim 0.1$M), AdapterGNN~\cite{li2024adaptergnn} ($\sim 0.1$M), and PHGNN ($\sim 0.03$M). The results show that PHGNN offers significant advantages in terms of parameter efficiency in addition to the performance improvement already illustrated in Table \ref{tab:prompt comparison}. Specifically, our approach only uses 6\% tunable parameters compared to fine-tuning. Note also that our approach outperforms both GPF and GPF-plus by a large margin while only introducing a very small overhead, as only 5\% more tunable parameters (wrt fine-tuning) are used compared with GPF.

\section{Conclusion}\label{sec:conclusion}
In this paper we introduced PHGNN, a novel framework for AD diagnosis and MCI conversion prognosis based on hypergraphs and prompt learning. Our framework leverages multiple modalities of imaging and tabular data and can cope with common scenarios in medical applications where labeled data is scarce. To achieve this, we introduced a novel prompt learning method tailored for hypergraphs. The resulting framework is shown to significantly outperform SOTA multimodal classification models, including  (hyper)graph based ones.


%
%
%
%
\bibliographystyle{splncs04} 
\bibliography{ref}
\end{document}